\documentclass[12pt]{article}

\usepackage[a4paper,margin=1in]{geometry}
\usepackage{setspace}
\usepackage{graphicx}
\usepackage{booktabs}
\usepackage{longtable}
\usepackage{array}
\usepackage{amsmath,amssymb}
\usepackage{natbib}
\usepackage[hidelinks]{hyperref}
\usepackage{enumitem}
\usepackage{authblk}
\usepackage{comment}
\setcitestyle{authoryear,round,semicolon,aysep={},yysep={;}}

\title{\textbf{InVitroVision: a Multi-Modal AI Model for Automated Description of Embryo Development using Natural Language}}

\author[1]{Nicklas Neu}
\author[2]{Thomas Ebner}
\author[3]{Jasmin Primus}
\author[1]{Raphael Zefferer}
\author[1]{Bernhard Schenkenfelder}
\author[3]{Mathias Brunbauer}
\author[1,*]{Florian Kromp}
\affil[1]{Software Competence Center Hagenberg GmbH,Softwarepark 32a, Hagenberg, 4232, Upper Austria, Austria}
\affil[2]{Kinderwunsch Zentrum Kepler Universitätsklinikum, Krankenhausstraße 26, Linz, 4021 Upper Austria, Austria}
\affil[3]{Wunschkind Klinik Dr. Brunbauer, Ebendorferstraße 6/4 Vienna, 1010, Vienna, Austria}

\affil[*]{Correspondence: florian.kromp@scch.at}

\date{}

\begin{document}
\maketitle


\section*{Abstract}

\textbf{Study question:}
Can foundational vision-language models be utilized to predict natural language descriptions of embryo morphology and development with limited resource availability?

\textbf{Summary answer:}
PaliGemma-2, a recent foundational multi-modal, vision-language model can be finetuned with only 1,000 images to enable the prediction of embryo morphology and development using natural language.

\textbf{What is known already:}
AI-based algorithms have been introduced to support experts' decisions within multiple parts of the IVF process. Although promising results have been achieved demonstrating the capability of AI models to improve consistency and standardization of decisions, these models rely on annotated data to be adapted to custom clinical data and tasks, a costly and tedious process. Foundational vision-languages can learn from multi-modal data to generate natural language description of images and can be finetuned to specific tasks with small datasets, but have not been introduced to the field of IVF yet.

\textbf{Study design, size, duration:}
1,100 frames of a publicly available embryo time lapse dataset have been annotated by clinical embryologists with natural language descriptions of morphology, embryonic cell cycle and developmental stage. A vision-language model was finetuned to learn the image captioning task.

\textbf{Participants/materials, setting, methods:}
1,000 annotated frames were used as training set, where subsets of 400, 700 and 1,000 images were used to train multiple models called InVitroVision, based on PaliGemma-2. A commercial model, ChatGPT 5.2 was used for comparison. A novel metric was developed to quantitatively assess the performance of the different models.

\textbf{Main results and the role of chance:}
Finetuned InVitroVision models, particularly the 1,000-sample variant, outperformed ChatGPT 5.2 and base models in overall accuracy. Performance consistently improved with larger training datasets, showing marked gains in developmental stage, embryonic cell cycle and morphological detail recognition, while ChatGPT 5.2 tended to over-specify unsupported detail.

\textbf{Limitations, reasons for caution:}
A single dataset was used to finetune the InVitroVision models, domain shift to images or frames from other microscope or incubator settings remain for future exploration.

\textbf{Wider implications of the findings:}
Foundational vision-language models are known to generalize to other modalities with limited data finetuning requirements. They are known to be excellent for few-shot adaptation to multiple other downstream tasks such as Gardner scoring. In addition, natural language descriptions of embryo morphology allow to use large language models to retrieve information and scientific evidence from a large set of relevant documents such as publications or consensus outcome.

\textbf{Study funding/competing interest(s):}
Supported by the Austria Wirtschaftsservice Gesellschaft mbH (aws), using funds from the National Foundation for Research, Technology and Development (Future Austria Fund) under the Grant No. P2508804. The authors declare no competing interests.

\textbf{Trial registration number:}
N/A.

\vspace{0.5em}
\textbf{Keywords:} Vision-language model, embryo morphology description, infertility, IVF, embryo, endometrium, reproductive medicine

\section*{Introduction}
In-vitro fertilization (IVF) has become an established treatment for infertility; however, success rates remain limited and are strongly influenced by the selection of embryos for transfer. A central component of IVF workflows is the morphological assessment of embryos, which is typically performed by embryologists using standardized grading systems based on visual inspection. Although these grading schemes provide valuable guidance, embryo evaluation remains inherently subjective and is affected by inter- and intra-observer variability. This variability in scoring has been reported for static images and in part for time-lapse sequences. \citep{Cimadomo2022} did use 80 static images of cleavage and blastocyst stage embryos to evaluate the intra- and inter-centre reliability in embryo grading performed according to the Istanbul Consensus (2015). It turned out that in general intra-centre reliability was better than between the different centres but it did not exceed beyond moderate or substantial. Upon interactive training sessions the overall reliability improved to moderate. There is also the idea that the agreement in scoring of static images will be worse in poor quality embryos as compared to good quality embryos \citep{Martinez2018}, in particular the presence of vacuoles, multinucleated blastomeres and the structure of the inner cell mass caused some inconsistency. Similarly, only fair to moderate agreement was observed for multinucleation and the evenness of blastomeres in a study using time-lapse videos \citep{Sundvall2013} while the more dynamic annotated parameters, such as pronuclear fading, cleavages up to 8-cell stage and hatching of the blastocyst resulted in extremely close agreement. Interestingly, the level of experience does not seem to play a major role in annotation \citep{Martinez2018}. 

It has been hypothesised that the establishment of external quality assessment programmes \citep{Martinez2018} and artificial intelligence technique might further improve the reliability of embryo selection, one of the key processes in IVF. These factors limit reproducibility and complicate standardization across clinics, posing challenges for both, clinical decision-making and research.

In recent years, artificial intelligence (AI) has emerged as a promising approach to support embryo assessment by enabling automated, data-driven analysis of embryo development. Deep learning models, particularly convolutional neural networks, have been applied to static \citep{Salih2025} and time-lapse (TL) \citep{LifeWhisperer} embryo images to automate morphological grading, predict implantation potential, and rank embryos for transfer \citep{DIMITRIADIS2022435}. Several studies have demonstrated high performance in embryo classification and outcome prediction. For example, \citep{Khosravi2019} developed the STORK model using time-lapse embryo images, achieving an area under the curve (AUC) of 0.987 for blastocyst quality classification and demonstrating generalization across clinical sites. Similarly, \citep{Wang2021} proposed a multichannel deep learning approach based on multifocal bright-field images, achieving an AUC of 0.936 and highlighting biologically relevant features such as the trophectoderm, inner cell mass, and zona pellucida through visualization techniques. In addition, \citep{Goyal2020} explored the use of AI for predicting live-birth outcomes using large-scale clinical datasets. Despite these advances, the clinical adoption of AI systems in IVF remains limited. 

A major limitation of existing AI approaches in IVF is their reliance on uni-modal data representations, commonly focusing on image-based features or morphokinetic parameters. In reality, however, clinical embryo assessment is inherently multi-modal, integrating visual inspection with descriptive language and expert reasoning. Recent developments in vision–language models (VLMs) enable the joint modeling of images and natural language, providing a potential pathway towards more interpretable and clinically aligned AI systems. In particular, foundational vision-language models can be trained to predict natural language descriptions of images, a task called image captioning, by learning to relate visual image features to linguistic semantics from natural language descriptions during supervised training. In addition to image captioning, such models are known to be efficient models for adaptation to other downstream tasks, tasks that they have not been trained for. Since their vision backbone (the vision encoder of the model) is pre-trained on a large number of images, dataset size requirements for finetuning are limited. Models such as CLIP \citep{radford2021learning} or BLIP \citep{li2022blip} have demonstrated strong performance in aligning visual and textual representations, enabling tasks such as image captioning, classification, and multi-modal reasoning. 

The adaptation of large pretrained models to clinical domains is challenged by limited data availability and the risk of model overfitting. To overcome these limitations, parameter-efficient fine-tuning methods such as Low-Rank Adaptation (LoRA) have been proposed, enabling selective modification of pretrained model architectures to allow parameter update without full retraining \citep{hu2022lowrank}. By introducing low-rank update matrices into attention and projection layers, LoRA allows efficient domain adaptation with reduced computational cost and improved generalization, making it particularly suitable for biomedical applications with constrained datasets. In the field of medical health care, multi-modal foundation models have shown promise for clinical report generation and diagnostic support across imaging modalities \citep{liu2025multi-modal}. However, the application of such approaches to reproductive medicine in general, and embryo assessment in particular, remains largely unexplored.

The aim of this study was to investigate the feasibility of applying vision–language models to embryo imaging data ranging from oocyte to blastocyst stage. Using parameter-efficient methods, we finetuned a pretrained multi-modal model, PaliGemma-2 \citep{steiner2024paligemma2familyversatile} with annotated embryo images from a time-lapse dataset. We evaluated the potential of multiple variants of the model (we call the resulting model InVitroVision to indicate its applicability in the IVF process) to generate clinically meaningful textual descriptions of embryo morphology. We then compared the generated  descriptions to descriptions generated by a commercial model, ChatGPT 5.2. We demonstrate that the InVitroVision (IVV) model trained on 1,000 images (IVV-1000), is capable of generating accurate captions, describing embryo morphology, embryonic cell cycle and developmental stage.  To the best of our knowledge, this is the first study conducted to explore the use of LoRA-adapted vision–language models for multi-modal embryo assessment in reproductive medicine.

\section*{Materials and methods}

\subsection*{Data collection}
\begin{figure}[h]
\centering
\includegraphics[width=\linewidth]{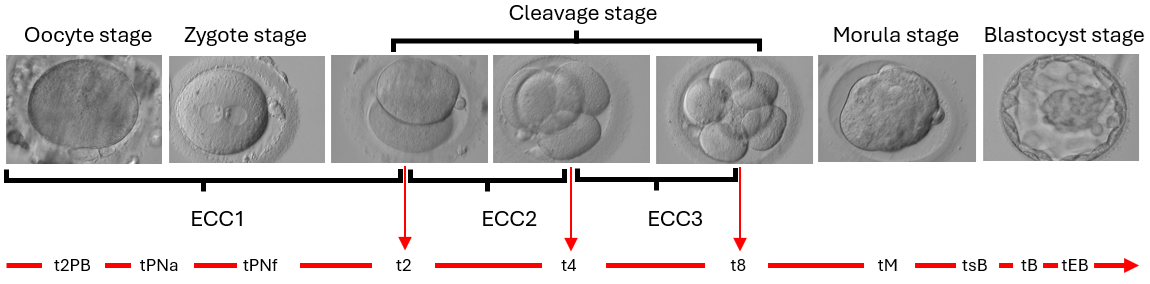}
\caption{Developmental stages, embryonic cell cycles and morphokinetic variables. 
}
\label{fig:cellcycles}
\end{figure}

To generate the training dataset used in this study, a subset of 1,100 images of a publicly available TL dataset of \citep{timelapseDS} was used without any modification. In other words, fotos were also used in case that the quality was suboptimal (e.g. cropped or dark). Single frames of oocytes and developing embryos were selected such that images from all developmental stages (oocyte, zygote, cleavage stage, morula, blastocyst) and embryonic cell cycles were represented (Figure \ref{fig:cellcycles}). ECCs were defined as suggested by \citep{ciray2014} reflecting the doubling from one to two (ECC1), two to four (ECC2) and from four to eight cells (ECC3). 

\subsection*{Data annotation} 
All frames were annotated by experienced embryologists by providing ECC and detailed morphological descriptions as ground truth (GT) description. The GT descriptions characterize oocytes (e.g. Figure \ref{fig:examples}, AAL839-6\_WELL6\_RUN027) as mature unfertilised gametes, embryos (e.g. Figure \ref{fig:examples}, AH988-4\_WELL4\_RUN236) as multicellular cleavage-stage or compacting embryos and blastocysts (e.g. Figure \ref{fig:examples}, AK383-1\_WELL1\_RUN425) as expanded, cavitated embryos with distinct cell lineages. In addition, morphological features are mentioned and, where applicable, their exact location is indicated. For example, the homogeneity of the cytoplasm and the zona pellucida were described as was the position of the polar bodies specified at oocyte stage. Embryos were mainly described based on blastomere count, fragmentation and degree of compaction. Any uncertainties regarding the exact blastomere number in case of moderate to heavy fragmentation were acknowleged. Furthermore, in the GT description blastocysts were graded according to expansion (which corresponds to the size of the blastocoel) as well as the quality of the inner cell mass and the trophectoderm.

\subsection*{Data split}
The dataset was split into a training- and a test set (training set: 1,000 images, test set: 100 images), such that images of all ECCs in both sets were equally distributed. 
The training set was further randomly split into multiple subsets (400 images and 700 images), to enable the evaluation of how training dataset size influences model performance. A comprehensive description of the dataset, including  annotation protocol, dataset split and validation strategy can be retrieved, see Section \textit{Data Availability}.

\subsection*{Model architecture and training}
To ensure consistent and high-quality model input, all images were resized to the required input resolution (448 × 448 pixels) while preserving the original aspect ratio through padding to avoid morphological deformation. To enable image captioning, a model-specific processor was used to perform image normalization, resizing, and text tokenization. The model relies on task-specific prompts to disambiguate caption generation from other possible multi-modal tasks.
For PaliGemma-2, each input starts with an $<$image$>$ token followed by the instruction “describe en”, prompting the model to generate an English-language description of the embryo.

To adapt the model to the domain of embryo images, LoRA adapters were utilized, allowing parameter-efficient finetuning with small datasets. While the pretrained backbone was not modified and its weights were frozen,  LoRA parameters were applied to both, its language and vision–language connector layers, and optimized during finetuning. This allows the model to better align domain-specific visual features with descriptive clinical terminology.

Training was performed using a supervised fine-tuning setup. PaliGemma-2 was optimized with AdamW, with batch sizes adjusted to available GPU memory. Gradient accumulation was applied to emulate larger batch sizes on limited hardware. During training, each model received the preprocessed embryo image in conjunction with the appropriate caption prompt and the corresponding ground-truth description. Performance was monitored on a held-out validation set to ensure stable convergence and prevent overfitting. Three models were finetuned: a model trained with a subset of 400 random images (InVitroVision-400), a model trained with a subset of 700 random images (InVitroVision-700), and a model trained with all (1,000) images of the training set (InVitroVision-1000). In addition to the IVV-models' performance, we report the performance of the non-finetuned model (Base).

\subsection*{ChatGPT comparison}
To investigate if a currently available large language models (LLM) from a commercial provider can process embryo images, we utilized ChatGPT 5.2, a recent LLM \citep{chatgpt}. We used the following prompt to force the model to provide accurate embryo descriptions: "Please describe the embryo morphology that you see and only if applicable, provide standard blastocyst grading terminology - if not applicable, do not state this. Describe compartments like inner cell mass and trophectoderm, zona pellucida if available - but if not applicable, because no blastocyst, don't describe this in the output (just present other morphological details or developmental stage or both). Output should be UTF-8 Decoding (not markdown or any other formating styles)". Given this prompt, we generated natural language descriptions for every image of the test set. 

\subsection*{Custom expert-based evaluation metric}

Automatic captioning metrics such as BLEU \citep{papineni-etal-2002-bleu}, ROUGE \citep{lin-2004-rouge}, METEOR \citep{banerjee-lavie-2005-meteor}, or CIDEr \citep{oliveira-dos-santos-etal-2021-cider} are not suitable for evaluating embryo image descriptions, as they primarily measure lexical overlap or semantic similarity but do not capture clinical correctness or completeness. In clinical embryology, errors such as incorrect identification of embryonic cell cycle are crucial \citep{ciray2014}, whereas linguistically different descriptions may still be clinically equivalent and correct. Furthermore, existing metrics cannot reliably distinguish between missing and incorrect information or assess morphological and spatial accuracy of morphological features. To this end, we developed a novel quantitative metric to rate the quality of models' predictions for single embryo static morphology images or TL frames. The score is based on four criteria: 

\begin{itemize}[noitemsep]
    \item \textbf{Embryo recognition (ER):} whether the presence of an embryo is explicitly identified by a model. ER is binary (0 = not recognized, 1 = recognized). Captions with $ER = 0$ are considered  incorrect and lead to an overall score of 0.
    
    \item \textbf{Embryonic cell cycle recognition (ECC):} correctness of identifying the embryonic cell cycle (e.g.\ oocyte, zygote, ECC2: 2-cell to 4-cell stage, ECC3: 4-cell to 8-cell stage, morula, blastocyst). ECCs according to \citep{ciray2014} are scored from 0 to 3, where 0 indicates an incorrect description, 1 no description of ECC, 2 a partially correct or ambiguous description of ECC (e.g. 6-cell instaed of 8-cell embryo), and 3 a correct identification of ECC.
    
    \item \textbf{Morphological details (MD):} correctness and completeness of morphological description. MD is scored from 0 to 4, with 0 indicating incorrect description of MD, 1 no description of MD, 2 partial correctness of MD, 3 largely correct MD with minor errors, and 4 entire correct MD.
    
    \item \textbf{Positional details (PD):} correctness of  spatial localization of relevant morphological features such as polar bodies, vacuoles, fragments or inner cell mass. PD is scored from 0 to 3, where 0 indicates incorrect positional description, 1 no positional description, 2 partially correct description, and 3 correct localization of positional details. If positional information was not applicable, PD was set to $-1$.
\end{itemize}

To obtain a single interpretable score while preserving clinical priorities, a weighted composite metric has been defined. If $ER = 0$, the overall score is set to 0.

Let $ECC_{\max} = 3$, $MD_{\max} = 4$, and $PD_{\max} = 3$.

If positional information is not applicable ($PD = -1$), the score is calculated as:

\begin{equation}
Score =
\frac{
\alpha \cdot \frac{ECC}{ECC_{\max}} +
\beta \cdot \frac{MD}{MD_{\max}}
}{
\alpha + \beta
}
\end{equation}

If positional information is available ($PD \geq 0$), the score is calculated as:

\begin{equation}
Score =
\frac{
\alpha \cdot \frac{ECC}{ECC_{\max}} +
\beta \cdot \frac{MD}{MD_{\max}} +
\gamma \cdot \frac{PD}{PD_{\max}}
}{
\alpha + \beta + \gamma
}
\end{equation}

Weights were experimentally determined and defined as $\alpha = 4$, $\beta = 2$, and $\gamma = 1$, reflecting clinical priorities by assigning the highest importance to embryonic cell cycle identification, followed by morphological accuracy and spatial localization of relevant morphological details. All components are normalized to ensure comparability across criteria while the score is bounded between 0 and 1.

\par To assess the quality of each prediction, generated captions were scored by $2$ experienced clinical embryologists by comparing predictions to the corresponding embryo images. Prior to the main evaluation, each embryologist independently scored predictions for $5$ captions, which were subsequently discussed to establish a consensus on the scoring criteria. For the full evaluation, each embryologist scored five model predictions for $50$ images, resulting in $250$ scores per embryologist. 

\section*{Results}

\subsection*{Expert-based evaluation}
The performance of different versions of the InVitroVision model (related to training set size) as compared to the base model and to Chat GPT 5.2 is listed in Table \ref{tab:results}.
\begin{table}[ht]
  \centering

  \begin{tabular}{|l|c|c|c|c|c|c|}
    \hline
    \textbf{Model} & \textbf{ER} & \textbf{ECC} & \textbf{MD} & \textbf{PD} & \textbf{Total score} &  \textbf{Total score PD}\\
    \hline
    BASE & 0.99 & 0.93 & 0.97 & 0.99 & 0.29 & 0.29\\
    \hline
    InVitroVision-400 & \textbf{1.00} & 2.04 & 1.89 & 0.98 & 0.57 & 0.59\\
    \hline
    InVitroVision-700 & \textbf{1.00} & 2.14 & 2.17 & 0.86 & 0.60 & 0.63\\
    \hline
    InVitroVision-1000 & \textbf{1.00} & \textbf{2.30} & \textbf{2.52} & \textbf{1.03} & \textbf{0.67} & \textbf{0.72}\\
    \hline
    ChatGPT & \textbf{1.00} & 1.87 & 1.62 & 0.98 & 0.52 & 0.55\\
    \hline
  \end{tabular}%
  \caption{Model performance of embryo image captioning on the test set evaluated using the developed metric. Best results per metric in bold. ER=Embryo recognition (max. score 1.0), ECC=Embryonic Cell Cycle (max. score 3.0), MD=Morphological details (max. score 4.0), PD=Positional details (max. score 3.0), Total Score (max. score 1.0), Total Score PD=total score restricted to cases where positional information was applicable (max. score 1.0).}
  \label{tab:results}
\end{table}

Embryo recognition (ER) scored high across all models. The base model achieved a score of 0.99, while all other models, including GPT-5.2 \citep{chatgpt}, achieved a score of 1.00, indicating reliable identification of the presence of an embryo across all approaches.

Substantial differences were observed in ECC. The base model showed limited performance (0.93), whereas finetuned models achieved markedly higher scores, increasing from 2.04 (400 samples) to 2.30 (1,000 samples). GPT-5.2 achieved a score of 1.87, outperforming the base model but remaining below all domain-adapted models.

A similar trend was observed for MD. The base model achieved a score of 0.97, while performance improved with increasing training data size, reaching 2.52 for the largest model. GPT-5.2 achieved a score of 1.62, again outperforming the base model but underperforming compared to all finetuned models.

Performance in PD showed smaller differences between models. The base model achieved a score of 0.99, comparable to the 400-sample model (0.98) and GPT-5.2 (0.98). The 700-sample model showed slightly reduced performance (0.86), whereas the 1,000-sample model achieved the highest score (1.03).

When restricting the analysis to cases where positional information was applicable, the 1,000-sample model performance improved markedly achieving the highest score (from 0.66 to 0.72). The performance of all other models only slightly increased, while the performance of the base model remained low.

\subsection*{Overall model performance}

Across all evaluation criteria, performance improved consistently with increasing training data size. The model trained on the largest dataset (1,000 samples) achieved the highest overall score (0.66), followed by the 700-sample model (0.60) and the 400-sample model (0.56). The pretrained foundation model (BASE) showed substantially lower performance (0.29). The general-purpose model (GPT-5.2) achieved an intermediate score (0.50), outperforming the base model but remaining below all domain-adapted models.

Overall, improvements were most pronounced for developmental stage recognition and morphological description, while gains in positional accuracy were more variable. These findings indicate that increasing amounts of domain-specific training data substantially enhance the clinical quality and consistency of generated embryo descriptions.

\subsection*{Performance related to embryonic cell cycle}
The result of model and training set size impact on evaluation results and related to embryonic cell cycle is listed in Table \ref{tab:results2}. 

\begin{table}[ht]
  \centering
  \resizebox{\textwidth}{!}{%
  \begin{tabular}{|l|c|c|c|c|c|}
    \hline
    \textbf{MV} & \textbf{Base} & \textbf{InVitroVision-400} & \textbf{InVitroVision-700} & \textbf{InVitroVision-1000} & \textbf{ChatGPT5.2} \\
    \hline
    tPB2 & 0.20$\pm$0.24 & 0.54$\pm$0.40 & \textbf{0.55$\pm$0.32} & 0.52$\pm$0.36 & 0.50$\pm$0.22 \\
    \hline
    tPNa & 0.16$\pm$0.16 & 0.64$\pm$0.30 & 0.60$\pm$0.27 & \textbf{0.65$\pm$0.30} & 0.51$\pm$0.25 \\
    \hline
    tPNf & 0.28$\pm$0.28 & 0.57$\pm$0.35 & 0.60$\pm$0.33 & \textbf{0.63$\pm$0.33} & 0.39$\pm$0.29 \\
    \hline
    t2 & 0.20$\pm$0.15 & 0.67$\pm$0.21 & 0.74$\pm$0.17 & \textbf{0.79$\pm$0.14} & 0.56$\pm$0.29 \\
    \hline
    t3 & 0.18$\pm$0.16 & 0.52$\pm$0.24 & 0.61$\pm$0.21 & \textbf{0.62$\pm$0.22} & 0.62$\pm$0.23 \\
    \hline
    t4 & 0.28$\pm$0.28 & \textbf{0.60$\pm$0.27} & 0.55$\pm$0.16 & 0.55$\pm$0.21 & 0.59$\pm$0.28 \\
    \hline
    t5 & 0.37$\pm$0.20 & 0.64$\pm$0.07 & 0.72$\pm$0.13 & \textbf{0.73$\pm$0.13} & 0.71$\pm$0.14 \\
    \hline
    t6 & 0.34$\pm$0.25 & 0.70$\pm$0.17 & 0.69$\pm$0.15 & \textbf{0.79$\pm$0.10} & 0.60$\pm$0.06 \\
    \hline
    t7 & 0.34$\pm$0.20 & 0.72$\pm$0.14 & 0.68$\pm$0.10 & \textbf{0.77$\pm$0.11} & 0.62$\pm$0.26 \\
    \hline
    t8 & 0.28$\pm$0.28 & 0.73$\pm$0.11 & 0.70$\pm$0.13 & \textbf{0.80$\pm$0.10} & 0.66$\pm$0.25 \\
    \hline
    t9+ & 0.40$\pm$0.23 & 0.46$\pm$0.21 & \textbf{0.67$\pm$0.24} & 0.63$\pm$0.25 & 0.66$\pm$0.12 \\
    \hline
    tM & 0.20$\pm$0.19 & 0.34$\pm$0.30 & 0.22$\pm$0.26 & \textbf{0.48$\pm$0.28} & 0.42$\pm$0.26 \\
    \hline
    tSB & 0.26$\pm$0.24 & 0.47$\pm$0.28 & 0.56$\pm$0.25 & \textbf{0.60$\pm$0.29} & 0.48$\pm$0.20 \\
    \hline
    tB & 0.47$\pm$0.28 & 0.46$\pm$0.35 & 0.63$\pm$0.36 & \textbf{0.69$\pm$0.19} & 0.18$\pm$0.32 \\
    \hline
    tEB & 0.45$\pm$0.31 & 0.50$\pm$0.24 & 0.55$\pm$0.12 & \textbf{0.76$\pm$0.10} & 0.29$\pm$0.31 \\
    \hline
  \end{tabular}%
  }
  \caption{Overall (total) model performance of embryo image captioning on the test set evaluated using the developed metric and related to embryonic cell cycle (ECC). ECC1=tPB2 to t2, ECC2=t2 to t4, ECC3=t4 to t8. Best metric marked in bold. MV=Morphokinetic variable.}
  \label{tab:results2}
\end{table}

Overall, the InVitroVision-1000 model performed best across all embryonic cell cycles, except for the 4-cell stadium (InVitroVision-400 performed best), the 9+-cell stadium (InVitroVision-700 performed best) and the time of second polar body extrusion stadium (InVitroVision-700 performed best). The base model overall performed worst across all embyonic cell cycles except for embryos reaching blastocyst stage, where the InVitroVision-400 model performed worst. ChatGPT5.2 could only compete at cleavage stage (e.g. ECCs 2 and 3). As soon as compaction started, results declined.

\subsection*{Qualitative Results}
\begin{figure}[p]
\centering
\includegraphics[width=\linewidth]{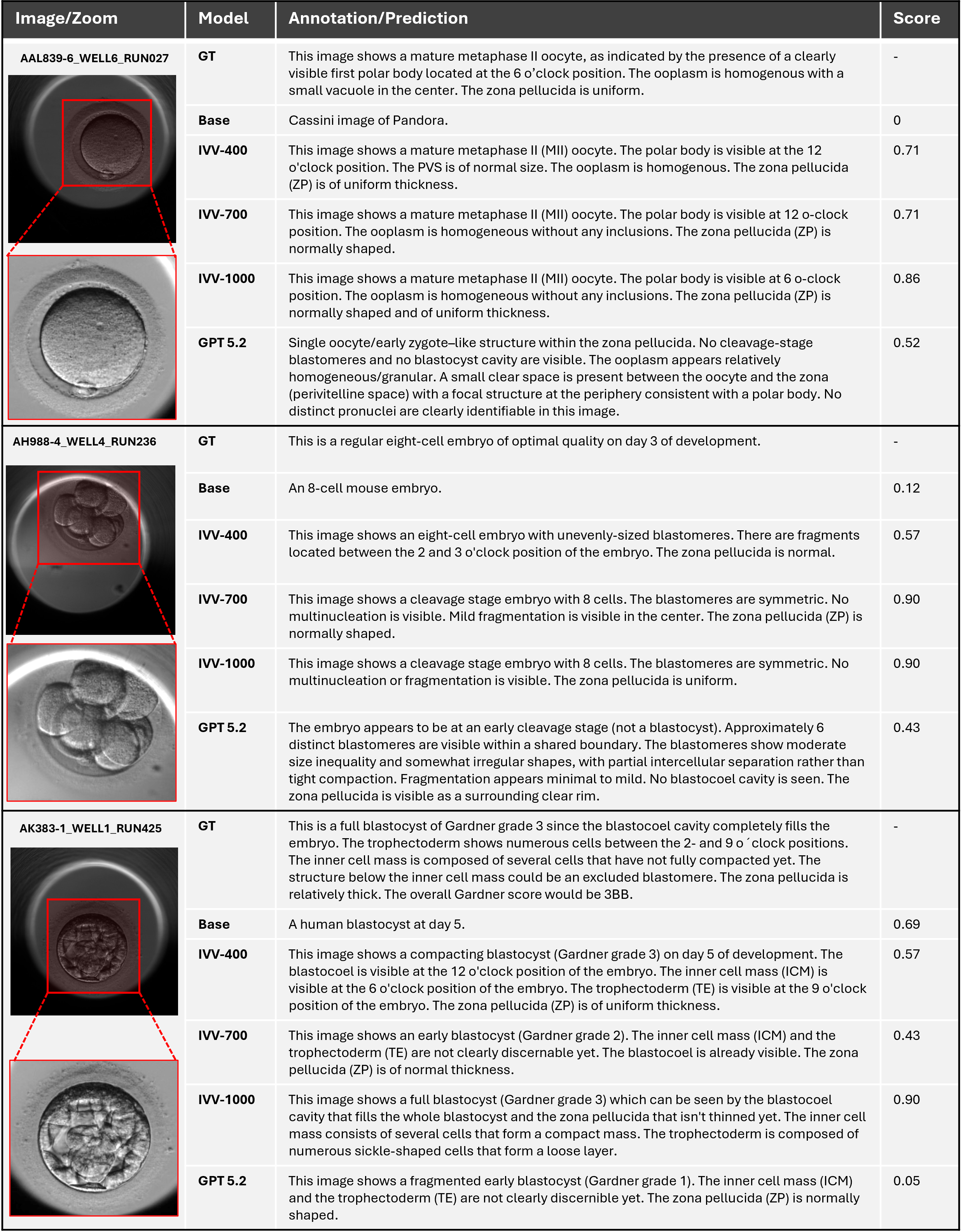}
\caption{Original examples of oocyte and embryo images, corresponding annotated captions (Groundtruth=GT) and evaluation scores. Zoom-in on relevant part of the images (increased brightness). IVV=InVitroVision. 
}
\label{fig:examples}
\end{figure}

The qualitative comparison of generated captions revealed clear differences in clinical accuracy and level of detail between models as shown in Table \ref{tab:results2} and Figure \ref{fig:examples}. The ground truth description characterizes the embryo as a multicellular cleavage-stage embryo with moderate fragmentation and a normal zona pellucida, while acknowledging uncertainty in exact blastomere count.

The base model generally performed the worst and in numerous cases didn't even identify an oocyte/embryo as such (e.g. AAL839-6\_WELL6\_RUN027). The IVV-400 model was relatively accurate in embryo staging but its score was often lowered because the suggested location of various morphological structures was incorrect (such as the location of the first polar body in AAL839-6\_WELL6\_RUN027). Model IVV-700 had its strenghts once embryos started to compact ($\geq$t9+) but the overall details were sometimes imprecise (e.g. no identification of inner cell mass in AK383-1\_WELL1\_RUN425). Overall, the IVV-1000 model demonstrated the best balance between accurateness in staging and identification of morphological details. In the rare cases location of certain morphological structures was provided (e.g. position of the polar body in AAL839-6\_WELL6\_RUN027), IVV-1000 was surprisingly accurate. Chat GPT-5.2, on the other hand, tended to use ,,unscientific language`` (e.g. using the term ,,within a shared boundary`` instead of ,,zona pellucida`` in AH988-4\_WELL4\_RUN236). Much more, some descriptions appeared to be inferred rather than directly supported by the image. Table \ref{tab:results2} indicates that the more blastomeres an embryo had and the more cells started to compact the worse became the performance of Chat GPT-5.2.

\section*{Discussion}

This study demonstrates the feasibility of using vision–language models to generate clinically relevant descriptions of embryo morphology from imaging data. In particular, models trained on increasing amounts of embryo-specific data showed consistent improvements in developmental stage recognition, morphological detail, and overall expert-based evaluation scores. While all models reliably detected the presence of an embryo (except for one image where the base model could not recognize an embryo), clinically relevant differences emerged in higher-level interpretative tasks. Notably, the largest finetuned model outperformed both the pretrained foundation model and the general-purpose language model (GPT-5.2), highlighting the importance of domain adaptation for specialized biomedical applications.

In principle, the general-purpose model (GPT-5.2) produced a fluent and structured description, in particular correctly identifying cleavage-stage embryos (ECC2, ECC3) also providing additional details such as approximate cell number and absence of multinucleation. However, performance regarding some aspects including the precise cell count and fragmentation level, indiactes a tendency towards over-specification. In contrast, the domain-adapted vision–language model (finetuned on 1,000 images) generated a more cautious and clinically grounded description. The model correctly identified the developmental stages, e.g. correctly distinguishing between compaction/morula and early blastocyst stage by explicitly noted the absence of a blastocoel cavity, supporting differentiation from a blastocyst. Furthermore, it described morphological features such as cytoplasmic granularity, partial compaction, and indistinct blastomere borders, while appropriately expressing uncertainty. These characteristics more closely align with expert reporting practices.

A key strength of this study is the use of an expert-based evaluation metric specifically designed for embryo description tasks. Unlike conventional natural language processing metrics, relying on lexical similarity, the proposed framework evaluates clinically relevant aspects such as morphological correctness and developmental stage identification. This enables a meaningful assessment of model performance in a clinical context. In addition, the study systematically investigates the effect of training data size, providing insights into how domain-specific data influences model behavior. The inclusion of both a pretrained foundation model and a general-purpose language model as baselines further strengthens the comparative analysis.

Several limitations must be considered. First, the study is based on a limited dataset derived from a specific clinical setting \citep{timelapseDS}, which may affect generalisability to other laboratories with different imaging systems or grading protocols. Second, although expert evaluation provides clinically meaningful assessment, it is inherently subjective and may be influenced by inter-observer variability. While efforts were made to standardize evaluation, residual variability cannot be excluded. Third, the study focuses on descriptive performance rather than direct clinical outcomes such as implantation or live birth, and therefore the clinical impact of the generated descriptions remains to be established. Finally, the evaluation of positional information showed variability across models, suggesting that spatial reasoning remains a challenging aspect for current approaches.

The observed performance improvements with increasing training data size suggest that the models benefit from learning domain-specific visual–textual associations. Fine-tuning likely enables the alignment of visual features with clinically meaningful terminology, improving both accuracy and consistency. The relatively strong performance of the general-purpose language model indicates that general medical or biological knowledge can partially transfer to this task. However, the consistent gap between GPT-5.2 and the finetuned models underscores the importance of specific training data. The variability observed in positional detail assessment may reflect limitations in the models’ ability to capture fine-grained spatial relationships, which are often subtle and context-dependent in embryo images.

From a clinical perspective, the ability to generate standardized, interpretable embryo descriptions has several potential implications. Such systems could support embryologists by improving consistency in reporting, facilitating documentation, and enabling more structured data collection across clinics. In research settings, standardized descriptions may enhance comparability between studies and support the development of large, harmonized datasets. Future work should focus on validating these models in multi-center settings, incorporating larger and more diverse datasets, and assessing their impact on clinical decision-making and outcomes. Thus, a potential future application of the proposed approach is the integration of generated embryo descriptions into retrieval-augmented generation (RAG) systems for clinical decision support. In such a framework, the vision–language model would first generate a structured, standardized textual description of embryo morphology. This description could then be used as a query to retrieve relevant information from a curated knowledge base containing clinical guidelines, embryology best practices, and evidence-based recommendations. Based on the retrieved information, the system could provide context-specific suggestions for clinical management, such as embryo selection strategies, prioritization for transfer, or recommendations for further observation. 
In addtion to context-specific text retrieval, multimodal foundation models relating language semantics to image features are known to be efficient for downstream tasks, by either finetuning models on limited data or by using them as few-shot learners (by providing only a few number of annotated samples to perform a task that the model has not been trained for). We demonstrated that  an early version of InVitroVision-1000 transforms embryo images into vector embeddings such that the shape of the resulting embedding space is a promising indicator for efficient downstream task training \citep{Kromp2026AIFoundationEmbryo}.

In conclusion, this study provides initial evidence that vision–language models, when adapted using parameter-efficient fine-tuning, can generate clinically meaningful descriptions of embryo morphology. These findings highlight the potential of multi-modal AI approaches to enhance interpretability and standardization in IVF workflows, while also underscoring the need for further validation in clinical practice.

\section*{Data availability}
The dataset used in this study is based on a publicly available dataset \citep{timelapseDS} and is publicly available \citep{Kromp2026fig}. A detailed description of the dataset can be retrieved elsewhere \citep{Kromp2026b}. More information about the research project behind this study is available \citep{birthai}. 

\section*{Authors' roles}
N. Neu performed all model trainings, experiments, and evaluations. R. Zefferer and B. Schenkenfelder created the tool to collect expert annotations. T. Ebner and J. Primus provided their expert knowledge by annotating the dataset and proofreading the manuscript. M. Brunbauer supported data annotation and proofread of the manuscript. F. Kromp coordinated the project, contributed in writing and proofreading of the manuscript. The AI-based Nature Research Assistant was used for proofreading and to make suggestions of possible improvements or how to shorten paragraphs. A self-hosted open-source large language model \citep{meta2025llama4scout} was used to suggest sentence rephrasing.

\section*{Funding}
Supported by the Austria Wirtschaftsservice Gesellschaft mbH (aws), using funds from the National Foundation for Research, Technology and Development (Future Austria Fund) under the Grant No. P2508804.

\section*{Conflict of interest}
The authors declare no conflict of interest.

\bibliographystyle{plainnat}
\bibliography{references}

\newpage

\end{document}